\documentclass[runningheads]{llncs}

\usepackage{hyperref}
\usepackage[utf8]{inputenc}
\usepackage[english]{babel}
\usepackage{amssymb}
\usepackage{amsmath}
\usepackage{latexsym}
\usepackage{amsfonts}
\usepackage{wrapfig} 
\usepackage{graphicx}
\usepackage{nccmath}
\usepackage{blkarray}
\usepackage{multirow}
\usepackage{times}
\usepackage{esint}
\usepackage{scalerel}
\usepackage{wrapfig}

\let\llncssubparagraph\subparagraph
\let\subparagraph\paragraph
\usepackage[compact]{titlesec}
\let\subparagraph\llncssubparagraph

 \usepackage[compact]{titlesec}
    \titlespacing{\section}{0pt}{3ex}{1ex}
    \titlespacing{\subsection}{0pt}{2ex}{1ex}
    \titlespacing{\subsubsection}{0pt}{0.5ex}{0ex}
    
\usepackage{setspace}
    {\end{pmatrix}\end{medsize}}%

\usepackage{tikz}
\usepackage{cite}
\usetikzlibrary{arrows,shapes,automata,backgrounds,petri,matrix,calc,patterns,fit}

\DeclareGraphicsExtensions{.pdf,.jpeg,.png}
\usepackage{graphicx}
\usepackage{caption}
\usepackage{subcaption}
\usepackage{enumerate}
\usepackage{float}
\usepackage{color,soul}
\usepackage{array,colortbl,xcolor}
\usetikzlibrary{tikzmark}
\usetikzlibrary{decorations.pathmorphing,shapes}

\usepackage[framemethod=tikz]{mdframed}
\usepackage{pdflscape}
\usepackage{todonotes}
\usepackage{algorithm}
\usepackage{algpseudocode}
\usepackage{adjustbox}
\usepackage{wrapfig,graphicx,lipsum}

\newcounter{sarrow}

\title{Predictive Business Process Monitoring via Generative Adversarial Nets: The Case of Next Event Prediction}
\author{Farbod Taymouri, Marcello La Rosa, Sarah Erfani, Zahra Dasht Bozorgi, Ilya Verenich}
\institute{
The University of Melbourne, Melbourne, Australia
\\
\email{\{farbod.taymouri, marcello.larosa,\\sara.erfani, ilya.verenich\}@unimelb.edu.au \\zdashtbozorg@student.unimelb.edu.au} 
}
\begin{document}
\maketitle
	
\begin{abstract}

Predictive process monitoring aims to predict future characteristics of an ongoing process case, such as case outcome or remaining timestamp. Recently, several predictive process monitoring methods based on deep learning such as Long Short-Term Memory or Convolutional Neural Network have been proposed to address the problem of next event prediction. However, due to insufficient training data or sub-optimal network configuration and architecture, these approaches do not generalize well the problem at hand. This paper proposes a novel adversarial training framework to address this shortcoming, based on an adaptation of Generative Adversarial Networks (GANs) to the realm of sequential temporal data.
The training works by putting one neural network against the other in a two-player game (hence the ``adversarial'' nature) which leads to predictions that are indistinguishable from the ground truth. We formally show that the worst-case accuracy of the proposed approach is at least equal to the accuracy achieved in non-adversarial settings. From the experimental evaluation it emerges that the approach systematically  outperforms all baselines both in terms of accuracy and earliness of the prediction, despite using a simple network architecture and a naive feature encoding. Moreover, the approach is more robust, as its accuracy is not affected by fluctuations over the case length. 
\end{abstract}


\section{Introduction}\label{sec: intorduction}
\noindent 
Predictive business process monitoring is an area of process mining that is concerned with predicting future characteristics of an ongoing process case \cite{TeinemaaDRM19, VerenichDRMT19}. Different machine learning techniques, and more recently deep learning methods, have been employed to deal with different prediction problems, such as outcome prediction \cite{TeinemaaDLM18}, remaining time prediction \cite{Tax17}, suffix prediction (i.e.\ predicting the most likely continuation of an ongoing case) \cite{Tax17,Lin2019MMPredAD,Camargo2019LearningAL}, or next event prediction \cite{EVERMANN2017129,Tax17,Lin2019MMPredAD,Camargo2019LearningAL,Pasquadibisceglie2019UsingCN}. 
In this paper, we are specifically interested in the latter problem: given an ongoing process case (proxied by a prefix of a complete case), and an event log of completed cases for the same business process, we want to predict the most likely next event by determining both its label (i.e.\ the name of the next process activity to be performed) and its timestamp (i.e.\ when such activity will start or complete). This problem has been addressed in \cite{EVERMANN2017129,Tax17,Lin2019MMPredAD,Camargo2019LearningAL} using Recurrent Neural Networks (RNNs) with Long-Short-Term Memory (LSTM), while \cite{Pasquadibisceglie2019UsingCN} uses Convolutional Neural Networks (CNNs) for predicting the next event label only. 

Despite their popularity, deep learning methods such as LSTM or CNN, often feature thousands to millions of parameters to estimate, and for this reason require lots of labeled training data to be able to generalize well the dataset at hand, as well as to learn salient patterns \cite{Goodfellow-et-al-2016}. In our context, this challenge is exacerbated by the limited size of real-life event logs available for training, compared to the number of parameters to be estimated. For example, an LSTM with one hidden layer containing 100 neurons has at least $4\times (100+1)^2$ parameters to be estimated, which in turn requires at least the same number of unique training instances, i.e.\ the same number of unique process cases in the event log. This is hardly the case in practice, as event logs typically contain several thousand or (at best) several million complete cases, of which only a subset are unique.

Motivated by Generative Adversarial Nets (GANs) \cite{GANNIPS2014_5423}, this paper proposes a novel adversarial training framework to address the problem of next event prediction. The framework is based on the establishment of a \textit{minmax game} between two players, each modeled via an RNN, such that each network's goal is to maximize its own outcome at the cost of minimizing the opponent's outcome. One network predicts the next event's label and timestamp, while the other network determines how realistic this prediction is. Training continues until the predictions are almost indistinguishable from the ground truth. During training, one player learns how to generate sequences of events close to the training sequences iteratively. Thus, it eliminates the need for a large set of ground truth sequences. 

To the best of our knowledge, this is the first paper that adapts GANs to the realm of temporal sequential data, for predictive process monitoring. This approach comes with several advantages. First, we formally show that the training complexity of the proposed adversarial net is of the same order as that of a net obtained via conventional (i.e.\ non-adversarial) training. Second, we show that the worst-case accuracy of our approach is not lower than that obtained via conventional training, meaning that the approach never underperforms a conventional approach such as LSTM with the same architecture. 

We instantiated our framework using a simple LSTM architecture for the two networks, and a naive one-hot encoding of the event labels in the log. Using this implementation, we evaluated the accuracy of our approach experimentally against three baselines targeted at the same prediction problem, using real-life event logs. 

The rest of this paper is organized as follows. The background and related work are provided in Sec. \ref{sec: backgorund and related work}. The presented approach is Sec. \ref{sec:proposed approach} while the evaluation is discussed in Sec.\ref{sec: evaluation}. Finally, Sec. \ref{sec: conclusion} concludes the paper and discusses opportunities for future work.

\section{Background and Related Work}
\label{sec: backgorund and related work}
\noindent In this section we provide background knowledge on machine learning with a focus on deep learning methods. Next, we discuss related work in predictive process monitoring, with a focus on next event prediction using deep learning.

\subsection{Machine learning and Deep Learning}
\label{subsec: ml}
\begin{wrapfigure}{r}{0.5\textwidth}
\vspace{-11mm}
  \begin{center}
    \includegraphics[width=0.55\textwidth]{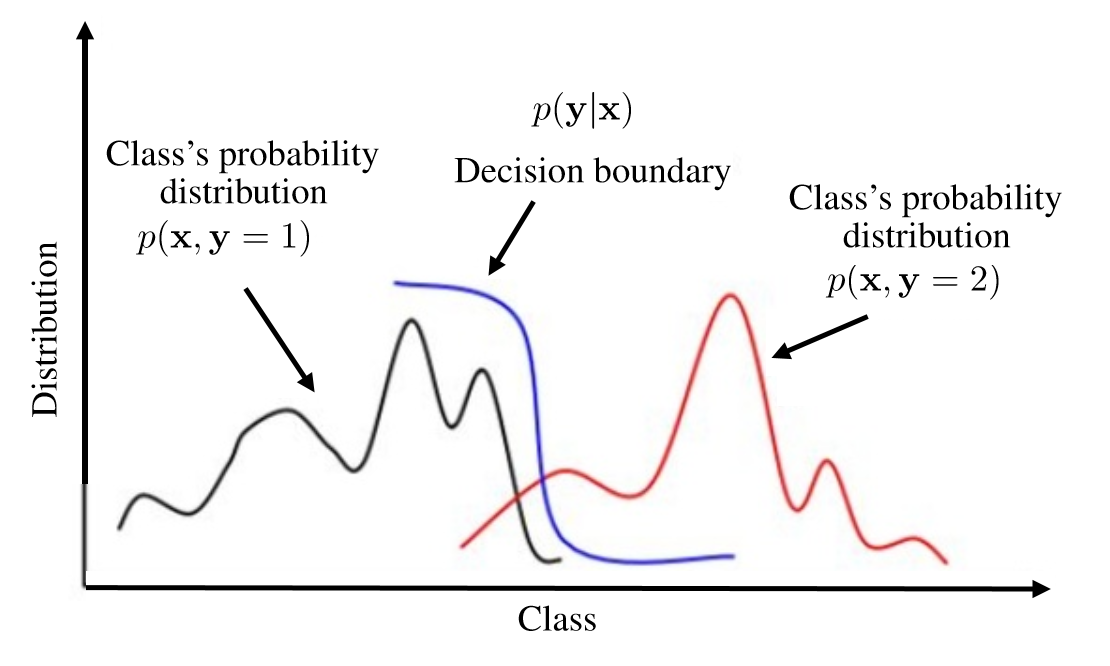}
  \end{center}
  \vspace{-6mm}
  \caption{Differences between a generative and discriminative models; $\mathbf{x}$ is the input's features, and $\mathbf{y}$ is the corresponding label}
  \vspace{-15pt}
  \label{fig:discGen}
\end{wrapfigure}

\noindent The goal of \emph{machine learning} is to develop methods that can automatically detect patterns in data, and  these patterns to predict future data or other outcomes of interest under uncertainty \cite{Murphy2012MachineL}. 
Depending on the underlying mechanisms, the learning model can be labelled as \emph{generative} or \emph{discriminative}. The objective of a generative model is to generate new data instances according to the given training set. In detail, it learns a joint probability distribution over the input's features.
The \emph{naive Bayes} classifier is an example of generative models.
In contrast, a discriminative model directly determines the label of an input instance by estimating a conditional probability for the labels given the input's features. 
\emph{Logistic regression} is an example of discriminative models. 
Discriminative models can only be used in supervised learning tasks, whereas generative models are employed in both supervised and unsupervised settings \cite{Ng2001OnDV}. Figure \ref{fig:discGen}, sketches the differences between the mentioned approaches; A discriminative model learns a decision boundary that separates the classes whereas a generative model learns the distribution that governs input data in each class.


\emph{Deep Neural Networks (DNNs)} are extremely powerful machine learning models that achieve excellent performance on difficult tasks such as speech recognition, machine translation, and visual object recognition \cite{Krizhevsky2012ImageNetCW, LeCun1998GradientbasedLA, Hinton2012DeepNN}. 
DNNs aim at learning feature hierarchies at multiple levels of abstraction that allow a system to learn complex functions mapping the input to the output directly from data, without depending completely on human-crafted features. The learning process in a DNN equals to estimating its parameters, and one can do it via \emph{Stochastic Gradient Descent (SGD)} or its modifications that are the dominant training algorithms for neural networks \cite{Goodfellow-et-al-2016}.



\emph{Recurrent Neural Networks (RNNs)} are a family of DNNs with \emph{cyclic} structures that make them suitable for processing sequential data \cite{BackProbRumelhart1986LearningRB}. RNNs exploit the notion of \emph{parameter sharing} that employs a single set of parameters for different parts of a model. Therefore, the model can be applied to examples of different forms
(different lengths) and generalize across them \cite{Goodfellow2013AnEI}. Such sharing is particularly important when a specific piece of information can occur at multiple positions within the input sequence. Two main issues in training an RNNs are \emph{catasrophic forgetting}, i.e., the model forgets the learned patterns, and \emph{optimization instability}, i.e., the optimization does not converge \cite{Goodfellow-et-al-2016}. The first issue can be alleviated by invoking the \emph{Long Short-Term Memory (LSTM)} architecture \cite{LSTM1997} which uses a few extra variables to control the information flow and thus causes the network to learn long-term patterns as well. The second issue can be mitigated by monitoring the gradient's norm of each parameter and scaling it down when it exceeds a threshold, a.k.a., \emph{gradient clipping} \cite{Pascanu2012OnTD}.



\begin{wrapfigure}{r}{0.6\textwidth}
\vspace{-11mm}
  \begin{center}
    \includegraphics[width=0.6\textwidth]{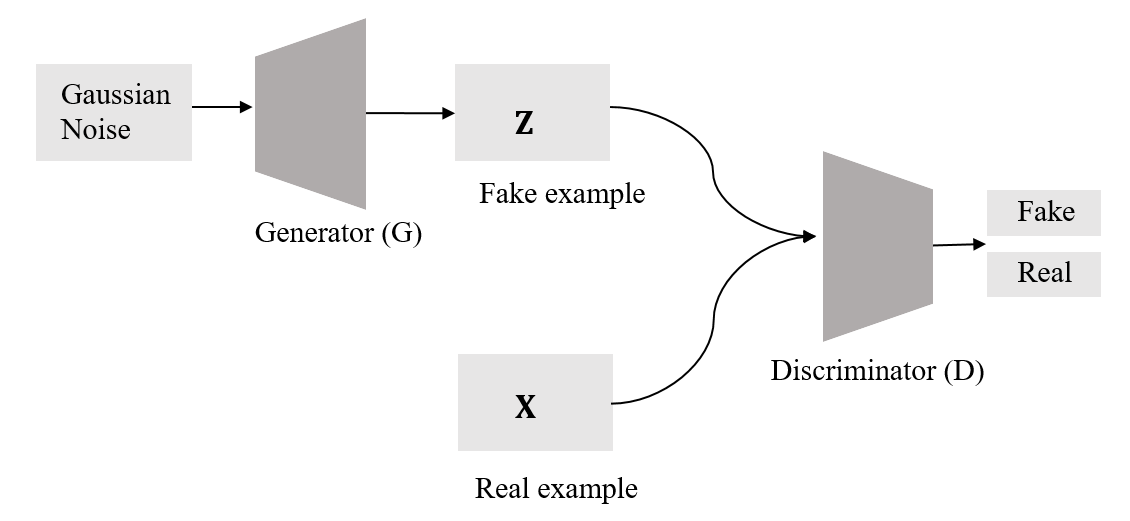}
  \end{center}
  \vspace{-6mm}
  \caption{Generative adversarial nets \cite{GANNIPS2014_5423}; the generator produces fake examples from Gaussian noise, and the discriminator determines which of its input is real or fake.}
  \vspace{-20pt}
  \label{fig:vanila gan}
\end{wrapfigure}

\noindent \emph{Generative Adversarial Nets (GANs)} \cite{GANNIPS2014_5423} is a framework that employs two neural network models, called players, simultaneously, see Fig. \ref{fig:vanila gan}.
The two players correspond to a \emph{generator} and a \emph{discriminator}. The generator takes \emph{Gaussian noise} to produce instances, i.e., \emph{fake instances}, which are similar to input instances, i.e., \emph{real instances}. The discriminator is a binary classifier such as logistic regression whose job is to distinguish real instances from generated instances, i.e., fake instances. The generator tries to create instances that are as realistic as possible; its job is to fool the discriminator, whereas the discriminator's job is to identify the fake instances irrespective of how well the generator tries to fool it. It is an adversarial game because each player wants to maximize its own outcome which results in minimization of the other player's outcome.  The game finishes when the players reach to \emph{Nash equilibrium} that determines the optimal solution. In the equilibrium point the discriminator is unable to distinguish between real and fake instances.

GANs provide enormous advantages compared to other strategies for training generative models. For instance, one can learn the input's joint probability even it is very sharp and degenerated, although, it needs accurate coordination of the players, i.e., neural nets, according to the problem at hand. Thus, depending on the input type, a GAN gives rise to a robust generative model that synthesizes high-quality images, texts, and sequences. Also, the GAN's discriminator can be viewed as a feature selection mechanism since it selects the most important features of its inputs to discriminate fake and real instances \cite{Goodfellow2017NIPS2T}.

\subsection{Predictive Process Monitoring of Next Event}
\label{subsec: related work}
\noindent This section reviews work on next event prediction using deep learning techniques. The interested reader can find an overview and comparative evaluation of different predictive process monitoring approaches in \cite{TeinemaaDRM19,VerenichDRMT19}.

The work by Evermann et al. \cite{EVERMANN2017129} uses the LSTM architecture for the next activity prediction of an ongoing trace, although the authors mention that one can  predict other attributes such as the event's duration time. It uses embedding techniques to represent categorical variables by high dimensional continuous vectors; it uses a two hidden layer LSTMs with one hundred epochs, the input's dimension varies according to the embedding representation, ten-fold cross-validation, and dropout for each cell is 0.2. 

Tax et al. \cite{Tax17} propose a similar architecture based on LSTMs. This work uses a one-hot vector encoding to represent categorical variables. Given an ongoing process execution, the approach predicts the next activity and its timestamp, and the remaining cycle time and suffix until the end of the process execution. Suffix prediction is made by next activity predictions iteratively. The proposed approach uses a variety of architectures. However, the best results are based on two hidden layers (shared and multi-task) LSTM with one hundred neurons in each layer for all the prediction tasks. Their results show that the proposed framework outperforms the technique in \cite{EVERMANN2017129}.

The work in \cite{Pasquadibisceglie2019UsingCN} uses a Convolutional Neural Network (CNN) for the next activity prediction task in a running process execution. The authors propose a data engineering schema to represent the spatial structure in a running case like a two-dimensional image. In experiments the approach starts with a prefix of length one and increases the prefix length during the training until the best accuracy can be obtained on the validation set. They use three convolutional and max-pooling layers with 32, 64, and 128 filters, respectively. The experiments show an improvement over \cite{Tax17, EVERMANN2017129}.

Camargo et al. \cite{Camargo2019LearningAL} employ a composition of LSTMs and feedforward layers to predict the next activity and its timestamp and the remaining cycle time and suffix for a running case. The approach uses embedding techniques similar to \cite{EVERMANN2017129} to learn continuous vectors for categorical variables and then use them for the prediction task via LSTMs. Similar to \cite{Tax17}, different settings such as "specialized", "shared categorical", and "full shared" architectures are considered in the experiments. Also, different configurations are considered randomly from a full search space of 972 combinations. The experiments show improvements over \cite{Tax17, EVERMANN2017129}, and for the next activity prediction task this approach sometimes outperforms that in \cite{Pasquadibisceglie2019UsingCN}.

Lin et al. \cite{Lin2019MMPredAD} propose an encoder-decoder framework based on LSTMs to predict the next activity and the suffix of an ongoing case. Unlike the previous approaches, it uses all available information in input log, i.e., both control-flow and performance attributes, for the prediction tasks. Random embedding is used for each event and its attribute. The encoder maps an input sequence into a set of high dimensional vectors and the decoder returns it back into new sequence that can be used for the prediction tasks. The experimental setup of this approach is different from \cite{EVERMANN2017129, Tax17, Camargo2019LearningAL, Pasquadibisceglie2019UsingCN}. Specifically, while the previous approaches aim to fit a predictive model for each prefix length, \cite{Lin2019MMPredAD} considers all possible prefix lengths at once during the training and testing phases. 

\section{Approach}
\label{sec:proposed approach}

\begin{figure}[h]
\vspace{-5mm}
	\centering
	\includegraphics[width=1\linewidth]{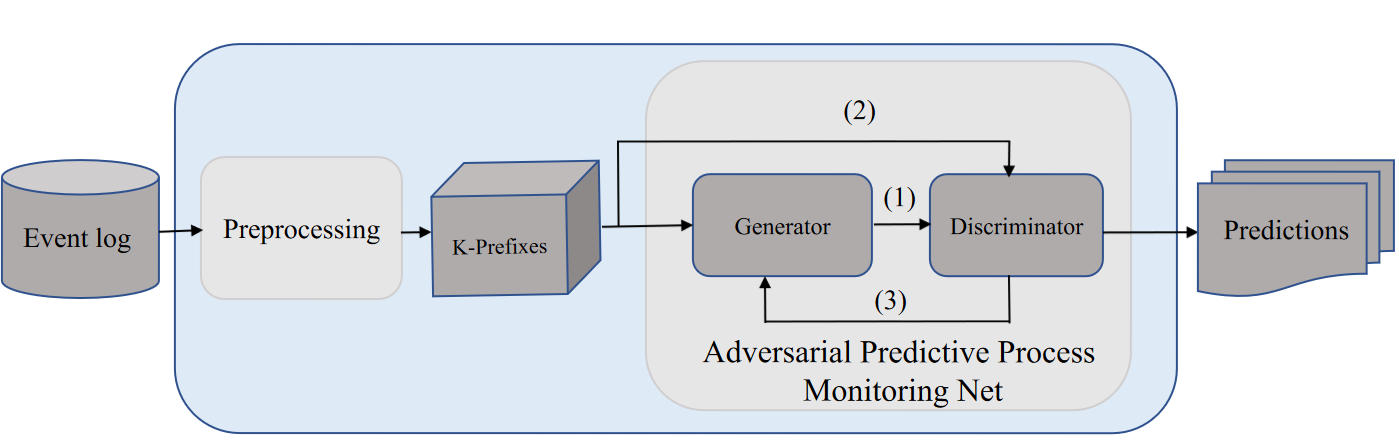}
	\caption{Overall approach for next event prediction}
	\label{fig:framework}
	\vspace{-5mm}
\end{figure}

\noindent The main aim of predictive process monitoring is to predict the corresponding attributes of ongoing process executions one or a few steps ahead of time. This paper, for an ongoing process execution (prefix), predicts an event's label and its timestamp one step ahead of time. To this end,  we propose an adversarial framework inspired by GANs \cite{GANNIPS2014_5423}, which coordinates players, i.e., the generator, and discriminator, in a novel way for process mining context, see Fig. \ref{fig:framework}. It has two main parts, \emph{data prepossessing}, and \emph{adversarial predictive process monitoring net}. The first part prepares the input data in the form of prefixes for the prediction task, and adopts the required encoding to deal with categorical variables. It uses one-hot encoding to manifest the viability of the proposed adversarial net. The second part establishes a minmax game between generator and discriminator by proposing fake and real prefixes. Real prefixes are those in the training set, and fake prefixes are formed from the generator's output, i.e., predictions.
The training runs as a game between two players, where the generator's goal is to maximize the accuracy of the prediction to fool the discriminator, and the discriminator's goal is to minimize its error in distinguishing real and fake prefixes, see flows (1), (2) in Fig. \ref{fig:framework}. It is an adversarial game since the generator and the discriminator compete with each other, i.e., learning from the opponent's mistake, see flows (1), (3) in Fig. \ref{fig:framework}. Thus maximizing one objective function minimizes the other one and vice versa.


The proposed adversarial net in this work has a number of major differences from the original GANs  proposed by Goodfellow et al. \cite{GANNIPS2014_5423}, i.e., Vanilla GAN, which are the core contributions of this paper. In our work, both the discriminator and the generator are composed of RNNs (LSTM architecture) and feedforward neural networks, rather than only feedforward networks. This is due to the fact that we apply GANs for sequential temporal data that LSTMs have been shown to perform well on \cite{LSTM1997}. Besides, the fake examples are formed by the generator's predictions and the input prefix; this is in contrast to Vanilla GAN that creates fake examples from Gaussian noise, see Fig. \ref{fig:vanila gan}. In this way, one can adopt GAN-like frameworks to the wide range of process mining applications. Finally, the proposed framework guarantees that, in the worst case, the generator has performance as if it was trained conventionally, i.e., no adversarial game, thus, it reduces the effects of \emph{mode collapse} in Vanilla GAN wherein the generator fails to model the distribution of the training data well enough, which in turn results in underfitting the input data and causes poor performance.

The rest of this section is organized as follows. First, the preliminary definitions are presented. Following that, we formalize the required data prepossessing. Next, RNNs training will be provided in detail, which will be used later in our framework. Finally, we give details of the adversarial predictive process monitoring net, including its training and optimization.

\subsection{Preliminaries and Definitions}
\label{sec: preliminaries}
\noindent This section provides the required preliminaries and definitions for the formalization of the proposed approach.

\begin{definition}[Vector]
\label{def:vector}
A vector, $\mathbf{x} = (x_1,x_2,\dots, x_n)^T$, is a column array of elements where 
the $ith$ element is shown by $x_i$. 
If each element is in $\mathbb{R}$ and vector contains $n$ elements, then the vector lies in $\mathbb{R}^{n \times 1}$, and the dimension of $\mathbf{x}$, $dim(\mathbf{x})$, is $n\times 1$.
\end{definition}

\noindent We represent a vector by a lowercase name in bold typeface. Beside, a set of $d$ vectors as $\mathbf{x}^{(1)}, \mathbf{x}^{(2)}, \dots, \mathbf{x}^{(d)}$, where $x^{(i)} \in \mathbb{R}^{n \times 1}$. Also, they can be represented by a matrix $\mathbf{M} = ( \mathbf{x}^{(1)}, \mathbf{x}^{(2)}, \dots, \mathbf{x}^{(d)} )$ where $\mathbf{M} \in \mathbb{R}^{n\times d}$. We denote the $ith$ row of a matrix by $\mathbf{M}_{i,:}$, and likewise the $ith$ column by $\mathbf{M}_{:,i}$. 

\begin{definition}[Gradient]
\label{def:grad}
For a function $f(\mathbf{x})$ with $f: \mathbb{R}^n \rightarrow \mathbb{R}$, the partial derivative $\frac{\partial}{\partial x_i} f(\mathbf{x})$ shows how $f$ changes as only variable $x_i$ increases at point $\mathbf{x}$. With that said, a vector containing all partial derivatives is called $gradient$, i.e., $\nabla_{\mathbf{x}} f(\mathbf{x}) = ( \frac{\partial}{\partial x_1} f(\mathbf{x}), \frac{\partial}{\partial x_2} f(\mathbf{x}),\dots,\frac{\partial}{\partial x_n} f(\mathbf{x}))^T$.
\end{definition}





\begin{definition}[Probability Distribution]
\label{def:prob distrib}
For a random variable (vector) $\mathbf{x} \in \mathbb{R}^n$, a probability distribution is a function that is defined as follow: $p: \mathbb{R}^n \rightarrow [0,1]$. Similarly, for two random variables $\mathbf{x} \in \mathbb{R}^n, \mathbf{y} \in \mathbb{R}^m$, a joint probability distribution is defined as: $p: \mathbb{R}^n \times \mathbb{R}^m  \rightarrow [0,1]$.
\end{definition}



\begin{definition}[Expectation, Kullback–Leibler (KL) Divergence]
\label{def:KL}
The expectation of a function $f(\mathbf{x})$ where the input vector $\mathbf{x}$ that has a probability distribution $p(\mathbf{x})$ is defined as:  $\mathbb{E}_{\mathbf{x}\sim p}[f(\mathbf{x})] = \oint p(\mathbf{x})f(\mathbf{x})d\mathbf{x}$. Given two probability distributions $p_1()$ and $p_2()$, KL divergence measures the dissimilarity between two distributions as follows: $D_{KL} (p_1 \parallel p_2) = \mathbb{E}_{\mathbf{x}\sim p_1}[\mathrm{log} p_1 (\mathbf{x}) - \mathrm{log} p_2 (\mathbf{x})]$.
\end{definition}

\noindent A similar concept to measure the dissimilarity between two distribution is the \emph{cross-entropy} and defined as $H(p_1,p_2) = -  \mathbb{E}_{\mathbf{x}\sim p_1}[ \mathrm{log} p_2 (\mathbf{x})]$.

\begin{definition}[Event, Trace, Event Log]
\label{def:trace event log}
An $event$ is a tuple $(a, c, t, (d_1, v_1), \ldots, (d_m, v_m))$ where $a$ is the activity name (label), $c$ is the case id, $t$ is the timestamp, and $(d_1, v_1) \ldots, (d_m, v_m)$ (where $m \geq 0$) are the event attributes (properties) and their associated values.
A $trace$ is a non-empty sequence $\sigma = \langle e_1,\ldots, e_{n} \rangle$ of events such that $\forall i,j \in \{1, \dots, n\} \; e_i{.}c = e_j{.}c$. An event log $L$ is a multiset  \{$\sigma_1, \ldots \sigma_n$\} of traces.
\end{definition}

\noindent A trace (process execution) also can be shown by a sequence of vectors, where a vector contains all or part of the information relating to an event, e.g., event's label and timestamp. Formally, $\sigma = \langle \mathbf{x}^{(1)},\mathbf{x}^{(2)}, \dots, \mathbf{x}^{(t)} \rangle$, where $\mathbf{x}^{(i)} \in \mathbb{R}^n$ is a vector, and the superscript shows the time-order upon which the events happened. 


\begin{definition}[k-Prefix (Shingle)]
\label{def: prefix}
Given a trace  $\sigma = \langle e_1,\ldots,e_{n} \rangle$, a $k$-$prefix$ is a non-empty sequence $\langle e_i,e_{i+1},\dots, e_{i+k-1} \rangle$, with $ i \in \{1,2,\dots, n-k+1\}$, which is obtained by sliding a window of size $k$ from the left to the right of $\sigma$.
\end{definition}

\noindent The above definition, a.k.a. \emph{k-gram}, holds when an input trace is shown by a sequence of vectors. For example, the set of 2-prefix for $\sigma = \langle \mathbf{x}^{(1)},\mathbf{x}^{(2)},\mathbf{x}^{(3)},\mathbf{x}^{(4)} \rangle$,  is $\{\langle \mathbf{x}^{(1)},\mathbf{x}^{(2)} \rangle,  \langle  \mathbf{x}^{(2)},\mathbf{x}^{(3)} \rangle , \langle \mathbf{x}^{(3)},\mathbf{x}^{(4)} \rangle \}$.  

\subsection{Data Preprocessing}
\label{subsec: data preprocessing}

\noindent This section elaborates on preparing $k$-prefixes which constitute the training and test set.
In detail, the approach in this paper learns a function that given a $k$-prefix, $\langle \mathbf{x}^{(1)},\mathbf{x}^{(2)}, \\\dots, \mathbf{x}^{(k)} \rangle$, returns a vector, $\mathbf{y}^{(k)}$, that can be viewed as the next attribute (property) prediction.  
For the sake of simplicity, we only predict the next \emph{activity} and its \emph{timestamp}, see Def. \ref{def:trace event log}. For the timestamp attribute, we consider the relative time between activities, calculated as the time elapsed between the timestamp of one event and the event's timestamp that happened one step before. However, without loss of generality, one can include the prediction of other attributes.

There are several methods in literature to encode and represent categorical variables. Unlike the techniques in \cite{Camargo2019LearningAL, EVERMANN2017129, Lin2019MMPredAD}, which learn embedding representations for categorical variables, this paper, uses \emph{one-hot encoding}. The reason to adopt this rudimentary encoding is to manifest that the viability of the presented approach owes to the adversarial architecture and not to the data engineering part. Indeed, one can integrate various embedding representations.

In a nutshell, the one-hot vector encoding of a categorical variable is a way to create a binary vector (except a single dimension which is one, the rest are zeros) for each value that it takes. Besides, we use $\langle \mathrm{EOS} \rangle $ to denote the end of a trace. Formally:

\begin{definition}[One-Hot Encoding]
\label{def: one hot encoding}
Given a universal set of activity names $\mathcal{E}$, including $\langle EOS \rangle$, and trace $\sigma$, one-hot encoding is a function, $f(\sigma, \mathcal{E})$, that maps $\sigma$ into a sequence of vectors $\langle \mathbf{x}^{(1)}, \mathbf{x}^{(2)}, \dots,  \mathbf{x}^{(|\sigma|)} \rangle$, where, \small{ $\mathbf{x}^{(i)} \in \{1\} \cup \{0\}^{\mathcal{E}-1}$, $\forall i \in \{1,2,\dots, |\sigma|\}$}.
\end{definition}
\noindent For example, given $\mathcal{E} = \{a_1, a_2, a_3, a_4, a_5, \langle \mathrm{EOS} \rangle\}$, and $\sigma = \langle a_1 ,a_3, a_4, \langle \mathrm{EOS} \rangle \rangle$. The one-hot vector encoding of $\sigma$ is the following sequence of vectors:
\footnotesize{\begin{equation*}
f(\sigma, \mathcal{E}) = \langle (\underbrace{1,0,0,0,0,0}_\text{$a_1$}), (\underbrace{0,0,1,0,0,0}_\text{$a_3$}), (\underbrace{0,0,0,1,0,0}_\text{$a_4$}), (\underbrace{0,0,0,0,0,1}_\text{$\langle EOS \rangle$})   \rangle  
\end{equation*}}
\normalsize
\noindent Furthermore, if $\mathbf{x}^{(i)}$ shows the one-hot vector of  $e_i$, then, one can \emph{augment} the former with the other attributes of the latter. In this paper, as mentioned already, we augment one-hot vectors with the time elapsed between the timestamp of one event and the event’s timestamp time that happened one step before.

\vspace{-18pt}
\begin{table}[]
    \centering
\scriptsize{\begin{tabular}{|r|l|l|l|}
\multicolumn{1}{c}{} & \multicolumn{2}{c}{$\overbrace{\rule{5cm}{0pt}}^{\mathbf{x}^{(t)}}$}  & \multicolumn{1}{c}{$\overbrace{\rule{3.2cm}{0pt}}^{\mathbf{y}^{(t)}}$}    \\
  \hline
  \textbf{Input $3$-prefix}  & \textbf{One-hot vector} & \textbf{Timestamp (s)} & \textbf{One-hot vector (next)}\\ 
  \hline
  $\langle (a_1$, 26/12/2019 00:30 AM), & $(1,0,0,0,0,0)$ & 0  & $(0,0,1,0,0,0)$ \\ \cline{2-4} 
  $(a_3$, 26/12/2019 01:02 AM), & $(0,0,1,0,0,0)$ & 1920 &  $(0,0,0,1,0,0)$ \\ \cline{2-4} 
  $(a_4$, 26/12/2019 01:18 AM), & $(0,0,0,1,0,0)$ & 960  & $(0,0,0,0,0,1)$ \\ \cline{2-4} 
  $(\langle EOS \rangle) \rangle$ & $(0,0,0,0,0,1)$ & 0  & $null$\\ \cline{2-4} 
  \hline
\end{tabular} }
\caption{Preprocessing of input $k$-prefix }
\label{table:example}
	\vspace{-9mm}
\end{table}

For each $k$-prefix, $\langle \mathbf{x}^{(1)}, \mathbf{x}^{(2)}, \dots, \mathbf{x}^{(k)}\rangle$, we couple another $k$-prefix $\langle \mathbf{y}^{(1)}, \mathbf{y}^{(2)}, \dots, \mathbf{y}^{(k)}\rangle$, where $\mathbf{y}^{(t)}$, $\forall t \in \{1,2,\dots,k\}$, is the next ground truth vector after visiting $\mathbf{x}^{(t)}$.
It is worth noting that, $\mathbf{x}^{(t)}$ and $\mathbf{y}^{(t)}$ might have different dimensions. For example, the former can be a one-hot vector, whereas the latter refers to the next activity's timestamp, which is scalar. A set of such paired $k$-prefixes is considered for training and test set. For the above example, Table \ref{table:example} shows the augmented vectors, i.e., $\mathbf{x}^{(t)}$, containing one-hot vectors and 
non-standardized events timestamps, as well as the respective next attribute, i.e., $\mathbf{y}^{(t)}$. The last row shows the end of prefix which is discarded for the since it does not provide useful information.

\subsection{Training Recurrent Neural Networks}
\label{subsec: RNN graphical model}
\noindent This section provides the training of RNNs in detail, which we will use it later in our proposed framework. For the ease of exposition, we present the training for the traditional RNN \cite{BackProbRumelhart1986LearningRB}, although, the concepts hold for any RNN architectures such as LSTM.



\begin{wrapfigure}{R}{0.55\textwidth}
\vspace{-11mm}
  \begin{center}
    \includegraphics[width=0.6\textwidth]{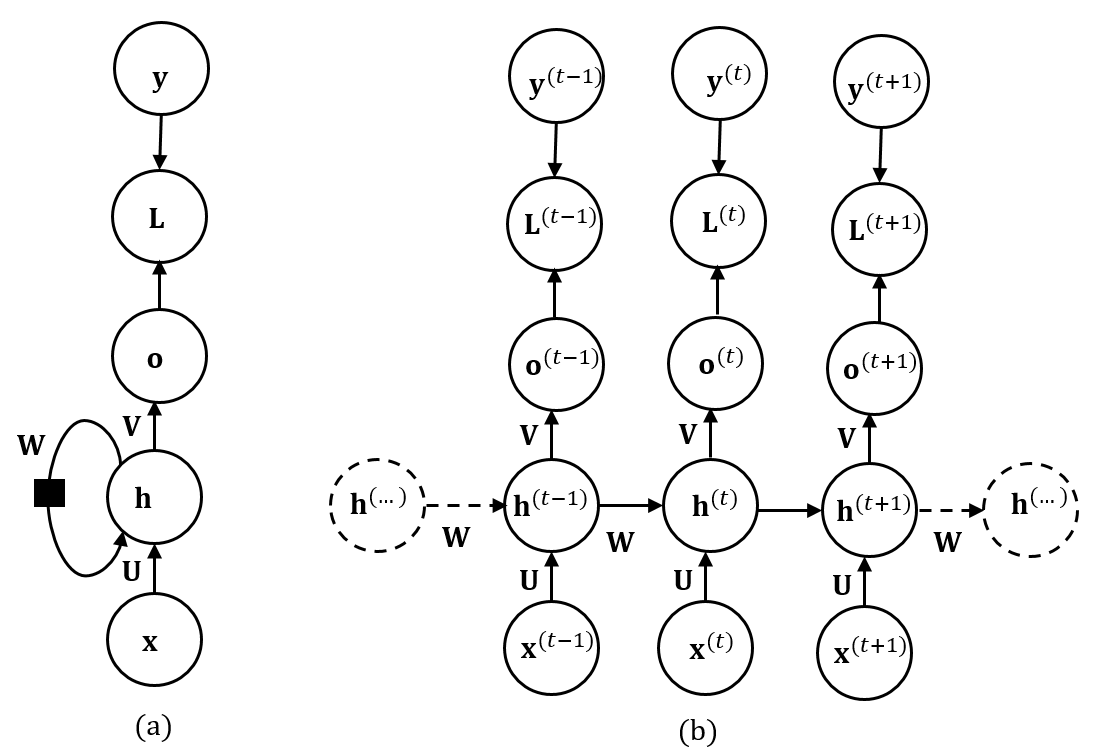}
  \end{center}
  \vspace{-5mm}
  \caption{(a) An RNN, (b) Time-unfolding of an RNN architecture}
  \vspace{-20pt}
  \label{fig:RNN}
\end{wrapfigure}

 Given a sequence of inputs $\langle \mathbf{x}^{(1)},\mathbf{x}^{(2)}, \dots, \mathbf{x}^{(k)} \rangle$, an RNN computes sequence of outputs $\langle \mathbf{o}^{(1)},\mathbf{o}^{(2)}, \dots, \mathbf{o}^{(k)} \rangle$ 
via the following recurrent equations:
\begin{equation}
\label{eq:rnn}
    \footnotesize{\mathbf{o}^{(t)} = \phi_o(\mathbf{V}^T\mathbf{h}^{(t)} + \mathbf{b}),\quad \mathbf{h}^{(t)} = \phi_h(\mathbf{W}^T\mathbf{h}^{(t-1)} + \mathbf{U}^T \mathbf{x}^{(t)}+\mathbf{c}) , \quad  \forall t \in \{1,2,\dots,k\} }
\end{equation}
\noindent Where $\mathbf{o}^{(t)}$ is the RNN's prediction for ground truth vector $\mathbf{y}^{(t)}$; $\phi_h$ and $\phi_o$ are nonlinear element-wise functions, and the set $\theta = \{\mathbf{W}, \mathbf{U}, \mathbf{V}, \mathbf{c, b}\}$, is the network's parameters. An RNN's architecture, and its time-unfolded graph are shown in Fig. \ref{fig:RNN} (a) and (b) respectively, where we hide vectors $\mathbf{c}$,$\mathbf{b}$, and functions $\phi_h$, $\phi_o$ for the purpose of transparency.

 One can estimate (learn) an RNN's parameters, i.e., $\mathbf{\theta}$, via the \emph{maximum likelihood principle}, in which $\mathbf{\theta}$ is estimated to maximize the likelihood of training instances. This way, an RNN is trained to estimate the conditional distribution of the next vector's attribute, $\mathbf{y}^{(t)}$, given the past input, $\mathbf{x}^{(1)}, \mathbf{x}^{(2)},\dots, \mathbf{x}^{(t)}$. In detail, to estimate $\mathbf{\theta}$, one minimizes the following loss function:
 \vspace{-3mm}
\begin{equation}
\vspace{-1mm}
\label{eq:mle}
    \small{J(\theta) = \sum \limits_{t=1}^{k} L^{(t)}, \quad \mathrm{where} \quad L^{(t)} = -\mathrm{log}\hspace{1mm} p_{m}(\mathbf{y}^{(t)}|\mathbf{x}^{(1)}, \mathbf{x}^{(2)},\dots, \mathbf{x}^{(t)} )}
\end{equation}

\noindent where $p_{m}$, gives the likelihood (probability) that the RNN generates the ground truth vectors. Besides, $L^{(t)}$, boils down to the cross-entropy between $softmax(\mathbf{o}^{(t)})$ and $\mathbf{y}^{(t)}$ whenever the latter is a one-hot vector, and it becomes $\parallel \mathbf{y}^{(t)} - \mathbf{o}^{(t)} \parallel_2$, a.k.a., Mean Square Error (MSE), for a continues ground truth vector. Finally, in an iterative way, the network's parameters are updated via SGD algorithm, wherein, the gradient of $J(\mathbf{\theta})$, i.e., $\nabla_{\theta} J$, is computed by \emph{backpropagation through time (BPTT)} \cite{BackProbRumelhart1986LearningRB}.

One can see that training an RNN or an LSTM in this way gives rise to a discriminative model, see Eq. \ref{eq:mle} and Fig. \ref{fig:discGen}. However, according to the \emph{Bayes' theorem} the estimated conditional distribution in Eq. \ref{eq:mle} is proportional to the joint probability distribution, i.e., $p_{m}(\mathbf{y}^{(k)}|\mathbf{x}^{(1)}, \mathbf{x}^{(2)},\dots, \mathbf{x}^{(k)} ) \propto p_{m}(\mathbf{x}^{(1)}, \mathbf{x}^{(2)},\dots, \mathbf{x}^{(k)}, \mathbf{y}^{(k)} )$. Thus, one can consider LSTMs or RNNs as generative models by using the learned distribution $p_m$ which is an approximation to the  input's ground truth joint distribution $p_d$. We will exploit this issue in our proposed framework.

\subsection{Adversarial Predictive Process Monitoring Nets}
\label{subsec: proposedGAn}
\noindent This section presents the core contribution of this paper by proposing an adversarial process to estimate a generative model for the predictive process monitoring tasks. The proposed framework is inspired by the seminal work in \cite{GANNIPS2014_5423}, i.e., Vanilla GAN, which has been used for synthesizing images. However, our proposed adversarial net is devised to work with time-series data, including categorical and continuous variables; Therefore, it is fully adaptable to a wide range of process mining applications.

\begin{figure}[h]
\vspace{-12mm}
	\centering
	\includegraphics[width=1\linewidth]{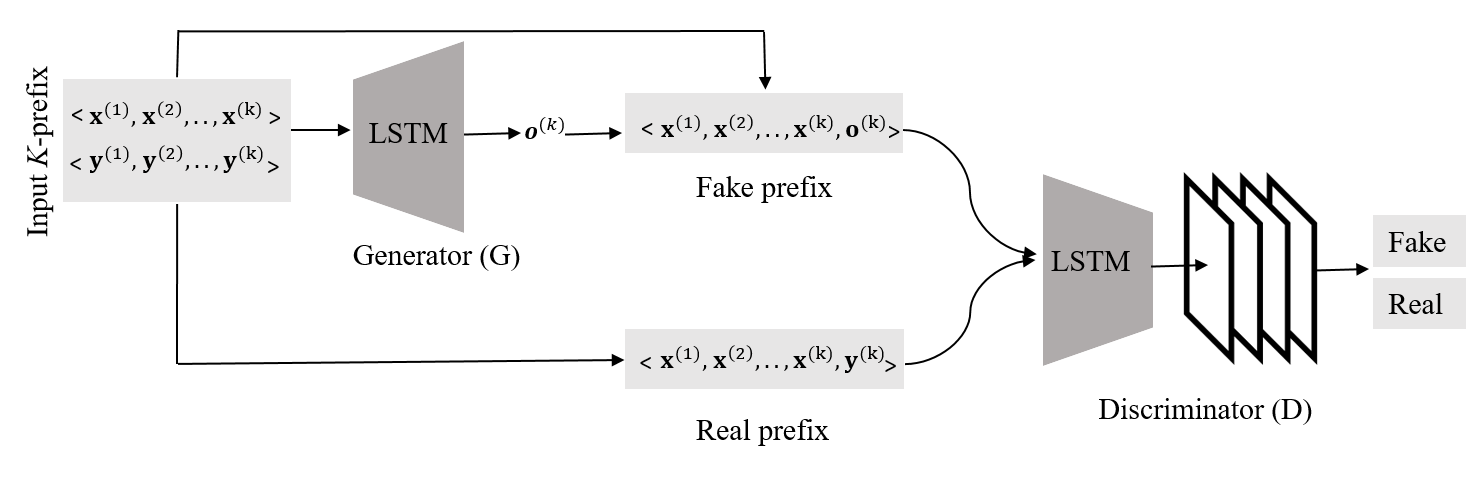}
	\vspace{-7mm}
	\caption{Proposed GAN architecture for predicting next attributes}
	\label{fig:gan}
	\vspace{-5mm}
\end{figure}

In the proposed adversarial architecture, shown in Fig. \ref{fig:gan}, both the generator and the discriminator are LSTMs, as explained in Sec. \ref{subsec: RNN graphical model}, and are denoted by $G(;\mathbf{\theta}_g)$, and $D(;\mathbf{\theta}_d)$ respectively. Precisely, the output of $G$ is a sequence, however, the last prediction is of our concern.
$D$ is equipped with an extra dense feedforward layer which assigns a probability to its input as a real prefix. 
The networks' parameters are denoted by $\mathbf{\theta}_g$ and $\mathbf{\theta}_d$, which are adjusted during training.


The generator in Fig. \ref{fig:gan} given a $k$-prefix $\langle \mathbf{x}^{(1)}, \mathbf{x}^{(2)},\dots, \mathbf{x}^{(k)} \rangle$ and its ground truth $\langle \mathbf{y}^{(1)}, \mathbf{y}^{(2)},\dots, \mathbf{y}^{(k)} \rangle$,
generates sequence $\langle \mathbf{o}^{(1)}, \mathbf{o}^{(2)},\dots, \mathbf{o}^{(k)} \rangle$ according to Eq. \ref{eq:rnn}. Thus, we want the $G$'s last prediction, $\mathbf{o}^{(k)}$, to be as close as possible to ground truth $\mathbf{y}^{(k)}$, such that, $D$ gets confused in discriminating
$\mathbf{o}^{(k)}$ and $\mathbf{y}^{(k)}$. To make this more concrete we define the followings
\emph{fake} and \emph{real} prefixes. 
\begin{equation}
\label{eq: fake example}
  \footnotesize{  \mathbf{X}^{(k)} = \langle \underbrace{\mathbf{x}^{(1)}, \mathbf{x}^{(2)},\dots, \mathbf{x}^{(k)},\mathbf{y}^{(k)}}_{\text{Real prefix}} \rangle, \quad \mathbf{Z}^{(k)} = \langle \underbrace{\mathbf{x}^{(1)}, \mathbf{x}^{(2)},\dots, \mathbf{x}^{(k)},\mathbf{o}^{(k)}}_{\text{Fake prefix}} \rangle        }
\end{equation}

\noindent Where  $\mathbf{X}^{(k)}$ and $\mathbf{Z}^{(k)}$ are sampled from $p_d$ and $p_m$ distributions respectively, and differ only in their the last elements. Thus, the minmax game, as an optimization,  is as follow:
\begin{equation}
\label{eq:minmax}
  \footnotesize{  \operatorname*{arg}\operatorname*{min}_{G} \operatorname*{max}_{D} =  \mathbb{E}_{\mathbf{X}^{(k)}\sim p_d}[\mathrm{log}\underbrace{D( \mathbf{X}^{(k)})}_{(a)}] + \mathbb{E}_{\mathbf{Z}^{(k)}\sim p_m}[\mathrm{log}\underbrace{(1-D(\mathbf{Z}^{(k)}))}_{(b)}]   }
\end{equation}

Equation \ref{eq:minmax} drives $D$ to maximize the probability of assigning $\mathbf{X}^{(k)}$ to a real prefix, see $(a)$, and  assigning $\mathbf{Z}^{(k)}$ to a fake prefix, see $(b)$. Simultaneously, it drives $G$ in generating fake prefixes, i.e., $\mathbf{Z}^{(k)}$s, to fool $D$ into believing its prefixes are real. In short, $G$ minimizes the cross-entropy between the ground truths and its predictions. Hence, the training procedure is presented in Alg. \ref{alg: ganTraining}.

\vspace{-6pt}
\begin{algorithm}[h]
	\caption{Stochastic gradient descent training of the proposed adversarial net}
	\label{alg: ganTraining}
\scriptsize{	\begin{algorithmic}[1]
	\For{ number of epochs } \Comment{{\fontsize{7}{5}\selectfont Number of training iterations}}
	\For{ each $\langle \mathbf{x}^{(1)}, \mathbf{x}^{(2)},\dots, \mathbf{x}^{(k)} \rangle$ } \Comment{{\fontsize{7}{5}\selectfont A $k$-prefix}}
	\State{\textbullet Generate $\langle \mathbf{o}^{(1)}, \mathbf{o}^{(2)},\dots, \mathbf{o}^{(k)}  \rangle$ using $G$ via Eq. \ref{eq:rnn}}
	
	\State{\textbullet Create fake and real prefixes, i.e., $\mathbf{Z}^{(k)}$ and $\mathbf{X}^{(k)}$, according to Eq. \ref{eq: fake example} }
	
	\State{\textbullet Update the discriminator, $D$, by \emph{ascending} its gradient: 
	\vspace{-4mm}
	\begin{equation*}
	\vspace{-4mm}
	   \mathbf{\theta}_d \leftarrow \mathbf{\theta}_d + \epsilon \bigg ( \nabla_{\mathbf{\theta}_d}[\mathrm{log}D(\mathbf{X}^{(k)}) + \mathrm{log}(1-D(\mathbf{Z}^{(k)}))] \bigg )
	\end{equation*}} 
	\State{\textbullet Update the generator, $G$, by \emph{descending} its gradient: 
	\vspace{-4mm}
	\begin{equation*}
	\vspace{-4mm}
	   \mathbf{\theta}_g \leftarrow \mathbf{\theta}_g - \epsilon \bigg ( \nabla_{\mathbf{\theta}_g}[\mathrm{log}(1-D(\mathbf{Z}^{(k)})) + J(\mathbf{\theta}_g)] \bigg)
	\end{equation*}} 
	\EndFor
	\EndFor
	\end{algorithmic}  }
\end{algorithm}
	\vspace{-5mm}

Algorithm \ref{alg: ganTraining} computes gradients for all prefixes in each epoch, although, one can use batches  to speed up training time as well. Besides, the learning rate, i.e., $\epsilon$, can be different for $G$ and $D$. Line 5 shows that the parameters of the discriminator, i.e., $\mathbf{\theta}_d$, are updated (maximizing) for each pair of real and fake prefixes by ascending the gradient of mistakes. Next, in line 6, we update (minimizing) the parameters of the generator, i.e., $\mathbf{\theta}_g$, by descending  the gradients of two terms. In the fist term, the generator exploits the discriminator's mistake in determining a fake prefix, i.e., $\mathbf{Z}^{(k)}$, to update its parameters (see flow (3) in Fig.~\ref{fig:framework}). This way, the generator learns how to fool the discriminator in the next iterations by generating more realistic prefixes. 
 The second term is the loss function as defined in Eq. \ref{eq:mle}. We incorporated this term because in some situations the $D$'s mistake for a fake prefix, i.e., $\mathrm{log}(1-D(\mathbf{Z}^{(k)}))$, does not provide sufficient gradient for $G$ to to update its weight. It happens at the beginning of training, when $D$ easily discriminates fake and real prefixes, e.g., $\mathrm{log}(1-D(\mathbf{Z}^{(k)})) = \mathrm{log}(1-0) = 0$, thus, adding $J(\mathbf{\theta}_g)$  facilitates the generator's learning process.\\

\noindent \textbf{Convergence}: At equilibrium, the generator’s prefixes, i.e., fake prefixes, are indistinguishable
from real prefixes, and it means that the generator has learnt the input data distribution, i.e., $p_d$. Thereby, its predictions must be enough close to ground truths. However, learning in GANs is a difficult task, since the minmax game in Eq. \ref{eq:minmax}, in general, is not a \emph{convex} function, thus, no global optimum solution is guaranteed to obtain. In addition, in a minmax game where each player reducing their own cost at the expense of the other player, reaching \emph{Nash equilibrium} is not guaranteed. Consequently, either of the mentioned issues causes GANs to underfit the input's data distribution which give rises to poor results \cite{GANNIPS2014_5423}. Alg. \ref{alg: ganTraining} alleviates the mentioned issues by invoking $J(\mathbf{\theta}_g)$ during training. Thus, in the worst case, the generator's ability to capture $p_d$ for the prediction task is lower bounded as if it was trained conventionally, i.e., no adversarial process.\\

\noindent \textbf{Complexity}: The complexity of Alg. \ref{alg: ganTraining} boils down to computing gradients for the generator and the discriminator. In detail, for  a $k$-prefix $\langle \mathbf{x}^{(1)}, \mathbf{x}^{(2)}, \dots, \mathbf{x}^{(k)}\rangle$ that is paired with $\langle \mathbf{y}^{(1)}, \mathbf{y}^{(2)}, \dots, \mathbf{y}^{(k)}\rangle$ , suppose that $\mathbf{x}^{(t)},\mathbf{y}^{(t)} \in \mathbb{R}^m$, $\forall t \in \{1,2,\dots,k\}$, and $\mathbf{U,W,V} \in \mathbb{R}^{m \times m}$.
    Therefore, to compute gradients of an RNN (or an LSTM architecture), one must do a forward propagation pass from left to right of the time-unfolded graph to generate $\langle \mathbf{o}^{(1)},\mathbf{o}^{(2)}, \dots, \mathbf{o}^{(k)} \rangle$ and to compute $J()$,  see Fig. \ref{fig:RNN} (b). Following that, a backward propagation pass moving right to left through the time-unfolded graph for computing gradients. In summary,  either a forward or a backward pass requires $O(km^2)$ operations \cite{Williams1995GradientbasedLA}. Thus, for a training set containing $n$ $k$-prefixes, $O(knm^2)$ operations are required in each iteration. Thereby, the  proposed adversarial net's complexity is of the same order as conventional training, i.e., no minmax game. Besides, it is noteworthy that the updates of the discriminator and the generator, i.e., lines 5 and 6, can be done in parallel after creating $\mathbf{Z}^{(k)}$ and $\mathbf{X}^{(k)}$.

\section{Evaluation}
\label{sec: evaluation}
\noindent We implemented our approach in Python 3.6 via PyTorch 1.2.0 and used this prototype tool to evaluate the approach over four real-life event logs, against three baselines \cite{Camargo2019LearningAL, Tax17, Pasquadibisceglie2019UsingCN}. The choice of the baselines was determined by the availability of a working tool, either publicly or via the authors. For this reason, we excluded from the experiment the work by Lin et al. \cite{Lin2019MMPredAD}, whose tool we were not able to obtain. Moreover, the work by Evermann et al. \cite{EVERMANN2017129} was excluded as Tax et al. \cite{Tax17} have already shown to outperform this approach.

The experiments were run on an Intel Core i8 CPU with 2.7 GHz, 64GB RAM, running MS Windows 10. The reason to use CPU rather than GPU is that the baselines were designed for CPU execution. However, our implantation also allows one to train discriminator and  generator on separate GPUs. Running of CPU instead of GPU only affects performance, not accuracy.


\subsection{Experimental Setup}
\noindent \textbf{Datasets:} The experiments were conducted using four publicly-available real-life logs obtained from the 4TU Centre for Research Data.\footnote{\url{https://data.4tu.nl/repository/collection:event_logs_real}} Table \ref{table:dataset} shows the characteristics of these logs while the description of the process covered is provided below. 

\begin{itemize}
\vspace{-2pt}
    \item \textbf{Helpdesk}: It contains traces from a ticketing management process of the help desk of an Italian software company. 
    \item \textbf{BPI12}: It contains traces from an application process for a personal loan or overdraft within a global financing organization. This process contains three sub-processes from which one of them is denoted as $W$ and used already in \cite{Tax17, Camargo2019LearningAL, EVERMANN2017129}. As such, we extract two logs from this dataset: BPI12 and BPI12(W).
    \item \textbf {BPI17}: It contains traces for a loan application process of a Dutch financial institute. The data contains all applications filed through an online system in 2016 and their subsequent events until February 1st 2017.
\end{itemize}

\begin{table}[h]
    \centering
    \vspace{-8mm}
\footnotesize{\begin{tabular}{|r|l|l|l|l|l|l|l|l|l|l|}
  \hline
  \textbf{Log}  & \textbf{Traces} & \textbf{Events} & \textbf{Labels} & \textbf{Max $|\sigma|$} & \textbf{Min $|\sigma|$} & \textbf{Avg $|\sigma|$} & \textbf{Avg $\Delta{t}$, days} & \textbf{St Dev($\Delta{t}$), days} \\ 
  \hline
  Helpdesk & 3,804     & 13,710 & 9  & 14 & 1 & 3.60 & 3.379 & 6.613 \\
  BPI12 & 13,087     & 262,200 & 23  & 175 & 3 & 20.03 & 0.453 & 1.719\\
  BPI12(W) & 9,658     & 72,413 & 6  & 74 & 1 & 7.49 & 1.754 & 3.075 \\
  BPI17 & 31,509     & 1,202,267 & 26  & 180 & 10 & 38.15 & 0.588 & 3.211\\
  \hline
\end{tabular} }
\caption{Descriptive statistics of the datasets ($|\sigma|$ is the trace length, $
\Delta{t}$ is the time difference between two consecutive event timestamps)}
\vspace{-10mm}
\label{table:dataset}
\end{table}

\noindent All the above logs feature event attributes capturing process resources. This information is used by the baseline in \cite{Camargo2019LearningAL} to extract extra signal for training.\\  

\noindent \textbf{Evaluation measures:} For consistency, we reuse the same evaluation measures adopted in the baselines. Specifically, to measure the accuracy of predicting the next event's label, we use the fraction of correct predictions over the total number of predictions. For the timestamp prediction, we report Mean Absolute Error (MAE), that is the average of absolute value between predictions and ground truths.\\

\noindent \textbf{Training setting:} For both generator and discriminator we use a two layer LSTM. In addition, the discriminator is equipped with a dense layer for the binary classification task. In detail:
\begin{itemize}
\vspace{-6pt}
    \item We use 25 epochs and split the data into 80\%--20\% for training and testing respectively, by preserving the temporal order between cases.
    However, early stopping is used to avoid over-training.
    In addition, we use a batch of size five to speed up the training procedure.
    \item For each log, we consider different prefix lengths to be used for the prediction task, i.e., $k$-prefixes, where $k \in \{2,4,6,8,10,15,20,25,30,35,40,45,50\}$, provided such $k$-lengths exist in the log. In detail, we train the proposed framework for each $k$-length case prefix and report the value of prediction accuracy and MAE, for that $k$. In this way, we can also observe the \textit{earliness} of a prediction, i.e.\ see how the predictions accuracy and MAE evolve over the prefix length. This experimental setup is in-line with that of \cite{Tax17, Camargo2019LearningAL, EVERMANN2017129, Pasquadibisceglie2019UsingCN}. Moreover, since the training and test set size varies for each prefix, we report the weighted average over all $k$-lengths for both accuracy and MAE.
    \item For each LSTM, we dynamically adjust the size of hidden units in each layer, and it is twice the input's size. For example if the augmented vectors dimension is 10, then each layer has $ 2\times 10$ hidden units.
    \item \emph{Adaptive Moment Estimation (ADAM)} is used as an optimization algorithm for both generator and discriminator. It accelerates the learning procedure by mitigating the effects of highly curvature search space \cite{Kingma2014AdamAM}. The learning rate, i.e., $\epsilon$, was set to $0.0002$ for both LSTMs to avoid gradient explosion during the training. In addition, we applied gradient clipping \cite{Pascanu2012OnTD} to scale down the gradient of each layer in every iteration. More specifically, let us use $\mathbf{g}$ to denote the gradient vector of a layer. Then, if $\frac{\|\mathbf{g}\|_2}{\mathrm{|batch|}}>10$, we scale the gradient as $\mathbf{g} = \frac{10\mathbf{g}}{\|\mathbf{g}\|_2}$. The threshold value of 10, only affects the learning speed and does not alter the learning outcome \cite{Pascanu2012OnTD}.
\end{itemize}

\noindent For the baselines, we used the best parameter settings, as discussed in the respective papers, or provided by the authors. These settings are provided in our tool distribution.

\begin{table}[h]
\vspace{-5mm}
	\centering 
	\footnotesize{ \begin{tabular}{|p{2.4cm}| p{1.2cm}| p{1.3cm} | p{1.0cm} | p{1.0cm} | p{1.2cm}| p{1.3cm}|p{1.0cm}|p{1.0cm}|}
		\hline
    \multicolumn{1}{|c|}{} & \multicolumn{4}{|c|} {\textbf{Weighted average accuracy}} & \multicolumn{4}{|c|}{\textbf{Weighted average MAE (days)}}\\
		\hline \hline
		\textbf{Approach} & Helpdesk  & BPI12(W) & BPI12 & BPI17 & Helpdesk  & BPI12(W) & BPI12 & BPI17\\ [1ex]
		\hline
		Ours & \textbf{0.9518}  & \textbf{0.9158}  & \textbf{0.9401}  & \textbf{0.9256}  & \textbf{0.8621}  & \textbf{0.6528} & \textbf{0.3471}  & 0.4225 \\
		Tax et al. \cite{Tax17} &0.7419  & 0.7077  & 0.7495  & 0.8941  & 3.660  & 1.5530 & 0.3716  & 0.5026 \\
		Camargo et al. \cite{Camargo2019LearningAL} &0.7384  & 0.7543  & 0.7182  & 0.8568  & 2.8996  & 1.8405 &0.5201  & \textbf{0.3646} \\
		Pasquadibisceglie et al. \cite{Pasquadibisceglie2019UsingCN} &0.7677  & 0.7734  & 0.7424  & 0.8676  & -  & - &-  & - \\
		\hline
	\end{tabular} }
		\caption{Weighted average accuracy for next label prediction, and Weighted average MAE for next timestamp prediction}
	\label{table:accuracy and MAE}
		\vspace{-\baselineskip}
\end{table}

\begin{figure}[h]
\vspace{-3mm}
	\centering
	\includegraphics[width=1\linewidth]{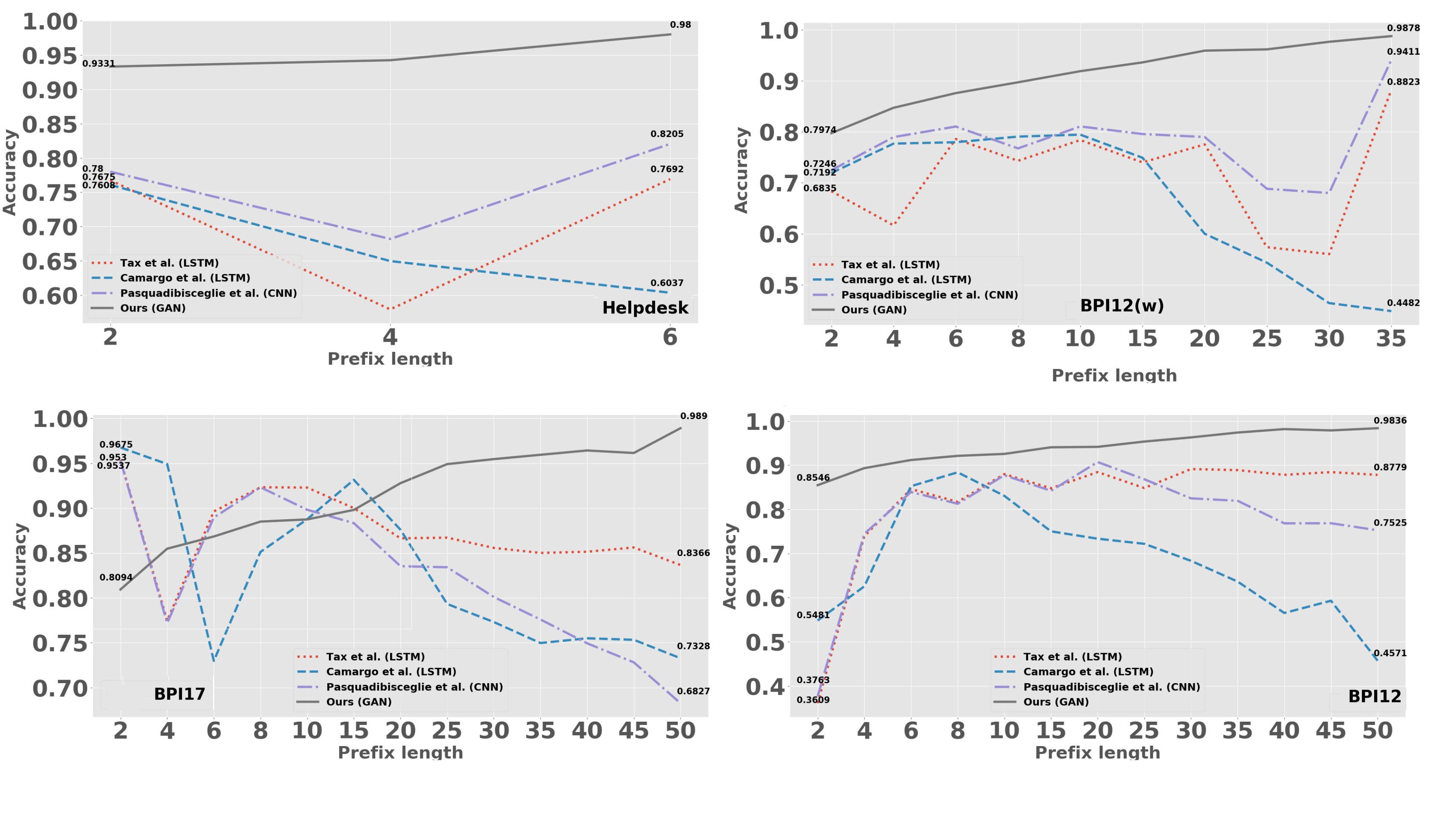}
	\vspace{-25pt}
	\caption{Accuracy of next event label prediction on the test set for different $k$-prefixes, $k \in \{2,4,\dots,50\}$; Our approach vs. baselines}
	\label{fig:acc pred}
	\vspace{-7mm}
\end{figure}

\subsection{Results}
\label{subsec:results}

\noindent\textbf{Next label prediction:} The second to fifth column of Table \ref{table:accuracy and MAE} show the weighted average accuracy of our approach and of the baselines, for each of the four logs. We can see that our approach provides a considerably more accurate overall prediction compared to the baselines for each dataset. In Figure \ref{fig:acc pred} we break this result for each $k$-prefix length, per log. From these charts we can draw several observations. First, our approach has an accuracy that is systematically higher than that of each baseline, at any given prefix length, obtaining at least $98\%$ accuracy for all logs at the longest considered prefix length. This is achieved by using a naive feature encoding (one-hot vector) of event labels, without extracting features from further event attributes such as resources. Second, the accuracy monotonically increases (though not strictly) with the length of the prefix. In contrast, the baselines exhibit fluctuations in accuracy as the length of the prefix increases. This is mainly due to the way a neural network is trained, and secondly, to the number of training examples (sequences of events in our case) used. In detail, our approach trains a neural network via a minmax game (adversarial) in addition to the conventional training, which allows us to obtain better generalization of the datasets at hand. Above that, the proposed approach is much less sensitive to the number of training sequences since the generator learns the input's distribution, through which it can then generate training sequences close to ground truth ones, thus eliminating the need for a large training data. The lack of sufficient training data severely impacts the the baselines. For example, \cite{Camargo2019LearningAL} loses accuracy faster than the other baselines as the prefix length increases. This is most likely because this approach extracts features from process resource, besides event labels and timestamps, and as such it requires a much larger training data for a larger number of parameters.

\begin{figure}[h]
\vspace{-9mm}
	\centering
	\includegraphics[width=1\linewidth]{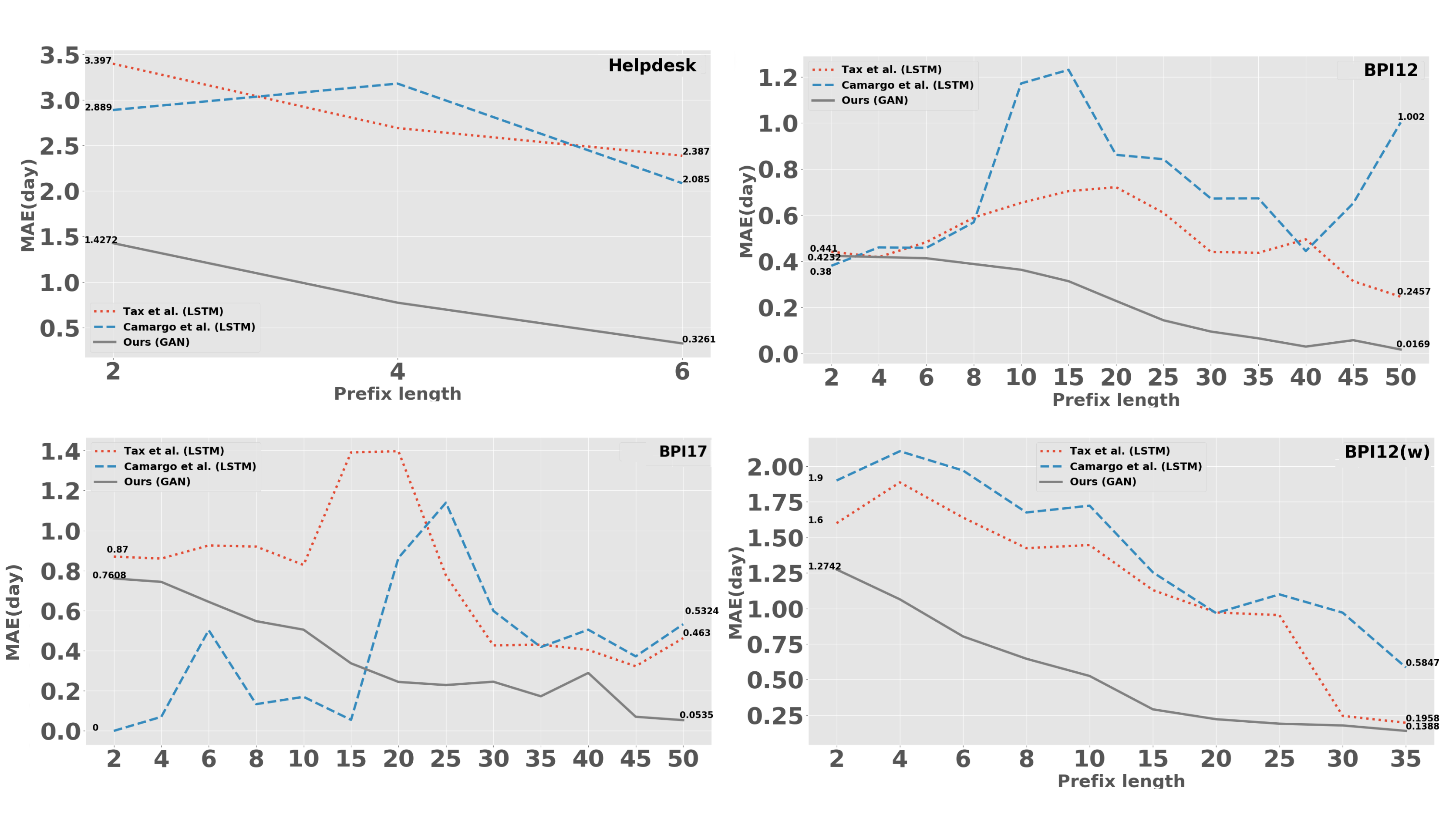}
	\vspace{-\baselineskip}
	\caption{MAE of next event timestamp prediction on the test set for different $k$-prefixes, $k \in \{2,4,\dots,50\}$; Our approach vs. baselines}
	\label{fig:mae timestamp}
	\vspace{-5mm}
\end{figure}

\noindent\textbf{Next timestamp prediction:} The last four columns in Table \ref{table:accuracy and MAE} show the weighted average MAE in days, for each log and for each approach, except \cite{Pasquadibisceglie2019UsingCN} as it does not support timestamp prediction. The detailed MAE for each prefix length in provided in Fig. \ref{fig:mae timestamp}. The results are consistent with those for next event label prediction, in terms of accuracy (lower error), earliness and stability. Specifically, from the charts we can see that for nearly all prefixes, our approach outperforms the baselines, except for $k=2$ in BPI12, BPI12(W) and for $k=2$--$15$ in BPI17, where \cite{Camargo2019LearningAL} provides slightly better MAE. For the BPI12 log, our approach reaches an MAE of 0.0169 at the longest prefix length, while the best result, achieved by \cite{Tax17} is 0.2457 (14 times higher). Given that MAE is measured as number of days, this means that there is an error of 14 days in the timestamp prediction. Looking at the weighted average MAE, we can observe the most significant improvements in the Helpdesk log, where our approach achieves up to 4 times lower MAE than the baselines. 

The higher MAE values of \cite{Camargo2019LearningAL} for certain prefix lengths, especially in the BPI17 log, are attributable to the use of resources in the log, and are in-line with the aggregate results in Table \ref{table:accuracy and MAE}, where \cite{Camargo2019LearningAL} outperforms our approach for the weighted average MAE in BPI17 (0.3646 instead of 0.4225). To confirm this intuition, we re-executed the experiment without using resources (we note that \cite{Camargo2019LearningAL} is the only baseline that extracts features from resources) and the accuracy obtained was lower (e.g.\ for BPI17, \cite{Camargo2019LearningAL} obtains a weighted average MAE of 0.5537 instead of 0.3646). 


In terms of stability, we can see that while we do not achieve monotonicity as in the case of next label prediction, the amplitude of the fluctuations of MAE in our approach is very small across all logs, with a clear downward trend as prefix length increases.\\

\begin{figure}[h]
\vspace{-10mm}
	\centering
	\includegraphics[width=1\linewidth]{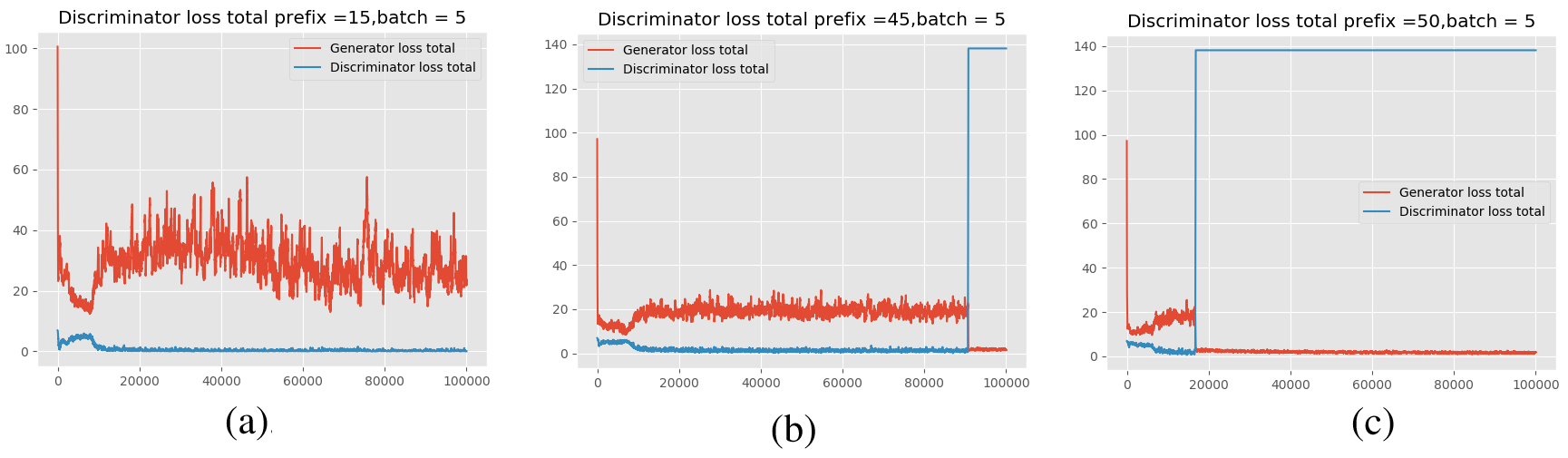}
	\caption{Convergence patterns based on loss functions of generator and  discriminator when training for BPI17: a) no convergence; b) late convergence; c) early Convergence}
	\label{fig:gan convergecne}
	\vspace{-24pt}
\end{figure}
\vspace{-\baselineskip}

\noindent\textbf{Behavior of the convergence:} We concluded our experiment by studying the convergence behavior of the generator and the discriminator while performing the minmax game in Alg. \ref{alg: ganTraining}. We provide three patterns that we observed in our experiments, as shown in Fig. \ref{fig:gan convergecne}, which plots the loss function of generator and discriminator. The patterns are the same for all datasets. As an example, we explain the pattern for the BPI17 log. Fig. \ref{fig:gan convergecne} (a) is an example where no convergence is made for this log. In other words, neither of the players can overcome the other. In this situation, the training continues with conventional training, as one can see from Fig. \ref{fig:acc pred} where our accuracy for BPI17 and for $k=15$ is slightly better than that in \cite{Tax17}. In contrast, Fig. \ref{fig:gan convergecne} (b) and (c) are examples of late, respectively, early convergence. Here the generator exploits the adversarial game since, after many iterations, it fools the discriminator as the discriminator's loss function increases significantly, and the generator's loss function drops. In such situations, the generator has learned the input's distribution correctly. Thus, the discriminator makes mistakes in distinguishing the ground truth from the generator's predictions. The effect of this gain can be seen in Fig. \ref{fig:acc pred} for BPI17 at $k=45$ or $50$, where our approach outperforms the baselines by far.

\section{Conclusion}
\label{sec: conclusion}
\noindent This paper put forward a novel adversarial framework for the prediction of next event label and timestamp, by adapting Generative Adversarial Nets to the realm of sequential temporal data. The training is achieved via a competition between two neural networks playing a minmax game. The generator maximizes its performance in providing accurate predictions, while the discriminator minimizes its error in determining which of the generator's outputs are ground-truth sequences. At convergence, the generator confuses the discriminator in its task. The training complexity of the proposed framework is of the same order as that of conventional training, and more importantly, we showed, both formally and empirically, that given the same network's architecture, our minmax training outperforms a network trained in conventional settings.

The results of the experimental evaluation highlight the merits of our approach, which systematically outperforms all the baselines, both in terms of accuracy and earliness. The results also show that the behavior of our approach is more robust as it does not suffer from accuracy fluctuations over the prefix length. This in turn confirms the generator's ability to learn the input distribution for generating predictions close to the ground truth, eliminating the need for a large number of training instances.

The experimental setting is limited to four (real-life) logs and three baselines. More extensive experiments should be conducted to confirm the results of this study. A further avenue for future work is to investigate alternative architectures within the proposed adversarial framework, to deal with other prediction problems such as case outcome or remaining time. More broadly, our adaptation of GANs to sequential temporal data lends itself well to various applications in process mining. For example, we foresee its use for variant analysis, automated process discovery, alignment computation in conformance checking, and process drift detection. We plan to investigate some of these opportunities in the future.

\smallskip\noindent\textbf{Reproducibility} The source code of our tool as well as the parameter settings used in our approach and in the baselines, in order to reproduce the experiments, can be found at \url{https://github.com/farbodtaymouri/GanPredictiveMonitoring}. This link also provides detailed experiment results.

\smallskip\noindent\textbf{Acknowledgments} We thank Manuel Camargo and Vincenzo Pasquadibisceglie for providing access and instructions to use their tools. This research is partly funded by the Australian Research Council (DP180102839).

\bibliographystyle{plain}
\bibliography{mybibfile}

\end{document}